\documentclass{article}

\usepackage{microtype}
\usepackage{graphicx}
\usepackage{subfigure}
\usepackage{booktabs} 
\usepackage[flushleft]{threeparttable}
\usepackage{authblk} 
\usepackage{hyperref}

\usepackage{amsmath}
\usepackage{amssymb}
\usepackage{mathtools}
\usepackage{amsthm}

\usepackage[capitalize,noabbrev]{cleveref}

\usepackage{natbib}
\theoremstyle{plain}

\theoremstyle{definition}

\theoremstyle{remark}

\newcommand{\varphivec}{\boldsymbol{\varphi}}

\title{Learning Linear Embeddings for Non-Linear Network Dynamics with Koopman Message Passing}

\author[*,1]{King Fai Yeh}
\author[1]{Paris Flood}
\author[2]{William Redman}
\author[1]{Pietro Liò}

\affil[*]{Corresponding Author: kfy22@cam.ac.uk}
\affil[1]{
Department of Computer Science and Technology,
University of Cambridge,
UK}
\affil[2]{
Interdepartmental Graduate Program in Dynamical Neuroscience,
University of California, Santa Barbara,
USA}

\date{}                

\begin{document}

\maketitle

\begin{abstract}
Recently, Koopman operator theory has become a powerful tool for developing linear representations of non-linear dynamical systems. However, existing data-driven applications of Koopman operator theory, including both traditional and deep learning approaches, perform poorly on non-linear network dynamics problems as they do not address the underlying geometric structure. In this paper we present a novel approach based on Koopman operator theory and message passing networks that finds a linear representation for the dynamical system which is globally valid at any time step. The linearisations found by our method produce predictions on a suite of network dynamics problems that are several orders of magnitude better than current state-of-the-art techniques. We also apply our approach to the highly non-linear training dynamics of neural network architectures, and obtain linear representations which can generate network parameters with comparable performance to networks trained by classical optimisers.

\end{abstract}

\section{Introduction}

Dynamical systems theory has long been a valuable branch of mathematics, finding applications in innumerable scientific fields. Systems fortunate enough to have a linear characterisation have well-developed methodologies for prediction, control, and interpretation, as their properties are defined by their spectral decomposition. In contrast, non-linear systems do not have a single, general theory which can determine their behaviour with universal precision. At best, there are bespoke solution methods for select scenarios - a frustrating state of affairs as non-linear phenomena are rich and ubiquitous. 

Koopman operator theory \cite{koopman_1931, mezic2005spectral, budivsic2012applied, brunton2022modern} has recently developed into an effective tool for casting dynamics problems from non-linear to linear. The Koopman operator progresses complex-valued measurements of a dynamical system linearly; unfortunately, the operator is also infinite-dimensional. As a remedy, many modern developments have focused on finding finite matrix approximations of the Koopman operator. This problem is challenging as it generally involves deriving potentially complex Koopman eigenfunctions.

Data-driven Koopman approaches have recently become extremely popular as they have demonstrated remarkable efficacy (see \cref{sec:background} for further details). However, current state-of-the-art methods do not perform well for non-linear dynamics with an underlying network geometry, despite sometimes being purposely designed for such a case. This is a deficiency that affects large classes of non-linear phenomena. 

Traditional approaches rely on in-depth analysis of a particular dynamical system; theoretical data-driven pipelines make general assumptions on the dynamics that impede their expressiveness. Existing deep learning approaches usually do not take in inductive biases corresponding to the underlying structures of particular dynamical systems.

In this paper, we present a novel, highly effective solution to this problem - a Koopman message-passing neural network (KMPNN). Our approach exploits the geometric structure of the dynamics when constructing a linear representation for the network. It leverages several message-passing layers to explore the neighbourhood of each node. These neighbourhood activations are then condensed into a single, global linear representation of the state of the dynamics. This layer enables control over the linear representation dimensions.

We run our approach on several network dynamical systems across a variety of scientific disciplines such as epidemiology (disease spread across social networks) and neuroscience (activation dynamics across the brain). Our approach produces a significantly better linear representation, through which we can predict dynamics trajectories several orders of magnitude more accurately than the state-of-the-art. 

To ramp up the complexity of the dynamics, we also evaluate our model on the training dynamics of a suite of neural network architectures. To the best of our knowledge, our approach again outperforms the best current methods at producing training trajectories, with a performance close to classical optimisers and a robustness to different architectural setups. On both types of dynamics, our model can find a linear representation from the start of the trajectories without the need for warm starts. Our results suggest that our KMPNN autoencoder can be used as a general framework for dissecting complex network dynamics due to its versatility across various levels of sophistication.

\section{Background}
\label{sec:background}

\subsection{Koopman Operator Theory}

\newcommand{\koopman}{\mathcal{K}_f}

Consider $\mathbf{x}_t\in \mathcal{X}\subseteq \mathbb{R}^n$ as the state of a dynamical system at time step $t$. In discrete-time dynamical systems, a dynamical map $f: \mathcal{X}\to \mathcal{X}$ is a function that advances the system state by one time step: $\mathbf{x}_{t+1} = f(\mathbf{x}_t)$. Consider a class of measurement functions $g\in \mathcal{G}: \mathcal{X}\to \mathbb{C}$ producing complex-valued measurements of the system at a given state. The Koopman operator $\koopman: \mathcal{G}\to \mathcal{G}$ for a dynamical map $f$ is a linear operator defined by $\koopman(g) \triangleq g\circ f$ \cite{koopman_1931}, satisfying:
\[\koopman(g)(\mathbf{x}_k) = g\circ f(\mathbf{x}_k) = g(\mathbf{x}_{k+1})\]
Since infinite degrees of freedom are needed to characterise the entire set of measurement functions, the Koopman operator is infinite-dimensional, constraining the practical use of the operator. Consider a Koopman-invariant subspace $\Hat{\mathcal{G}}$ spanned by a set of measurement functions $\{g_i\}^m_{i=1}$, such that $\koopman(g)\in \Hat{\mathcal{G}}$ for any $g\in \Hat{\mathcal{G}}$. By constraining the domain and co-domain of $\koopman$ to this invariant subspace, the Koopman operator can be represented with a finite matrix $\mathbf{K}\in \mathbb{C}^{m\times m}$ instead.

Consider a finite vector $\mathbf{g}\in \mathcal{X}\to \mathbb{C}^m$ of measurement functions $\mathbf{g} = \begin{bmatrix} g_1 & g_2 & \hdots & g_m \end{bmatrix}^{\intercal}$ with $g_i\in \Hat{\mathcal{G}}$. Applying $\mathbf{g}$ to a state $\mathbf{x}_k$ produces an embedding of measurements in $\mathbb{C}^m$, satisfying:
\[\mathbf{g}(\mathbf{x}_{k+t}) \triangleq \koopman^t(\mathbf{g})(\mathbf{x}_k) = \mathbf{K}^t(\mathbf{g}(\mathbf{x}_k))\]
If $\mathbf{g}$ is invertible, we have $\mathbf{x}_{k+t} = \mathbf{g}^{-1}(\mathbf{K}^t(\mathbf{g}(\mathbf{x}_k)))$, where $\mathbf{g}(\mathbf{x})$ effectively gives a linear representation to the dynamical states.

A Koopman eigenfunction $\varphi: \mathcal{X}\to \mathbb{C}$ is a particular type of measurement function satisfying $\koopman(\varphi) = \lambda\cdot \varphi$. If $\mathbf{g}$ is constructed as a vector of eigenfunctions $\varphivec$, the corresponding $\mathbf{K}$ will be diagonal, filled with their eigenvalues. Finding Koopman eigenfunctions allows for a decomposition of the action of the Koopman operator \cite{mezic2005spectral}:
\[\koopman(g)(\mathbf{x}_k) = g\circ f(\mathbf{x}_k) = \sum_{i = 1}^m \lambda_i \varphi_i v_i\]
where $v_i \in \mathbb{C}$ are Koopman modes. Such a decomposition is referred to as the Koopman mode decomposition. It illustrates that the complex dynamics can now be understood as evolving along exponential paths in the directions defined by $\varphi_i v_i$. Additionally, the time scales over which different dynamic phenomena exist can be assessed via the magnitude of the $\lambda_i$. This makes Koopman mode decomposition a powerful tool for analysis and prediction and has driven its use in many applied problems \cite{budivsic2012applied, mauroy2020koopman, brunton2022modern}.

\subsection{Related Work}

Finding a basis for measurement functions for a Koopman-invariant subspace is usually not straightforward. Traditional approaches attempt to construct a basis with expert knowledge of the targeted dynamical system. Modern Koopman analysis uses data-driven methods to approximate such a subspace, including theoretical pipelines \cite{dmd_2014, williams_2015, kaiser_2018, fujii_2019}, and more recent deep learning approaches \cite{lusch_kutz_brunton_2018, li_2019, otto2019linearly, yeung2019learning, azencot2020forecasting, han_2020, shi_2022}.

There have been several applications of Koopman analysis to the training dynamics of deep neural networks. Dogra established a connection between neural network training and dynamical systems via Koopman operator theory \cite{dogra2020dynamical}. Researchers have then leveraged the relationship to accelerate the training process \cite{dogra_redman_2020, tano_2020, simanek_2022, luo2022koopman}, by utilising the learned linear embeddings to predict the evolution of network weights and to prune deep neural networks \cite{manojlovic_2020, redman_2021}. 

The methods presented in this section have shown impressive results on various non-linear phenomena. However, as we shall see in \cref{sec:experiments}, even the state-of-the-art approaches leave much to be desired when applied to dynamics problems with network geometry. 

\subsection{Graph Neural Networks}

Graph neural networks (GNN) are a class of neural networks designed for processing data with graph structures. GNNs utilise message passing \cite{gilmer_2017} to aggregate the local neighbourhood features for each node. For a graph $G = (V, E)$ with node features $\mathbf{x}_u\in \mathbb{R}^c$ and edge features $\mathbf{e}_{u\to v}\in \mathbb{R}^d$, a message passing iteration is represented as:
\[\mathbf{x}_v' = \gamma(\mathbf{x}_v, \square_{u\to v}\ \phi(\mathbf{x}_v, \mathbf{x}_u, \mathbf{e}_{u\to v}))\]
$\phi$ generates a message on each edge $u\to v$ using its features and the features of its two endpoints. $\square$ is a permutation-invariant aggregator (e.g. sum, max) that combines the messages of a neighbourhood. $\gamma$ takes the aggregated message and updates the node in question. 

Different flavours of GNNs are created by varying the flexibility of the message generation function $\phi$ and node update function $\gamma$ \cite{kipf_2016, velickovic_2017, gilmer_2017, shi_2020}. In the most generic settings, usually called message-passing neural networks (MPNN), $\phi$ and $\gamma$ are fully-connected multi-layer perceptrons (MLPs) \cite{gilmer_2017}.

\subsection{Network Dynamics}

\begin{table*}[t]
\caption{Theoretical network dynamical systems from an ensemble of scientific disciplines including genetics, neuroscience, epidemiology, and ecology \cite{bakhtiarnia_2020}.}
\label{table:synthetic_network_dynamics}
\vskip 0.1in
\begin{center}
\begin{small}
\begin{sc}
\resizebox{\columnwidth}{!}{\begin{tabular}{lll}
\toprule
\textbf{Network Dynamics}   & \textbf{Equation}  & \textbf{Examples} \\
\midrule
Regulatory      & $\frac{dx_i}{dt} = -x_i^{0.4} + \sum_{j\to i} A_{ij} \frac{x_j^{0.2}}{1+x_j^{0.2}}$   & Gene regulation    \\[2ex]
Neuronal        & $\frac{dx_i}{dt} = -Bx_i + C\tanh(x_i) \sum_{j\to i} A_{ij} \tanh(x_j)$               & Activation dynamics between brain regions    \\[2ex]
Population      & $\frac{dx_i}{dt} = -x_i^{0.5} + \sum_{j\to i} A_{ij} x_j^{0.2}$                       & Birth-death processes    \\[2ex]
Epidemic        & $\frac{dx_i}{dt} = -x_i + \sum_{j\to i} A_{ij} (1 - x_i)x_j$                          & Social networks, Spread of epidemics    \\[2ex]
Mutualistic     & $\frac{dx_i}{dt} = x_i(1-x_i^2) + \sum_{j\to i} A_{ij} \frac{x_i x_j}{1+x_j}$         & Plant-pollinator relationships in ecology    \\
\bottomrule
\end{tabular}}
\end{sc}
\end{small}
\end{center}
\vskip -0.1in
\end{table*}

Complex dynamical networks are geometric structures that can display non-linear behaviour over time \cite{bakhtiarnia_2020}. Each node $u$ in the graph is associated with a numerical quantity $\mathbf{x}_u$. This quantity is governed by a specific differential equation concerning the quantities of its local neighbourhood, usually in the form of:
\[\frac{dx_u}{dt} = f(\mathbf{x}_u) + \sum_{v\to u} A_{vu}\cdot g(\mathbf{x}_u, \mathbf{x}_v)\]
where $f$ and $g$ represent self-dynamics and neighbourhood impacts. $A_{vu}$ is an adjacency matrix storing the edge weight of $v\to u$. If the network is unweighted, $A_{vu} = 1$ for any edge $v\to u$. If the network is undirected, $A_{vu} = A_{uv}$. Otherwise, if no edge exists between $v$ and $u$, $A_{vu} = 0$. \cref{table:synthetic_network_dynamics} shows several complex networks commonly appearing in biological, neural, and social dynamics.

\section{Approach}

This section presents a novel approach to learning a set of measurement functions giving non-linear network dynamics a linear representation. In line with the autoencoder developed by \citet{lusch_kutz_brunton_2018}, our method learns a set of Koopman eigenfunctions that produce measurements of the network which correspond one-to-one with the actual network states. To exploit the intrinsic geometric structures, we present the KMPNN autoencoder to improve the quality of the linear representations obtained.

\subsection{Problem Formulation}

We hope to learn a set of Koopman eigenfunctions $\varphivec$, their inverses $\varphivec^{-1}$ and a diagonal matrix $\mathbf{K}$ satisfying:
\[\mathbf{x}_{k+t} = \varphivec^{-1} (\mathbf{K}^t (\varphivec(\mathbf{x}_k)))\]
where $\mathbf{x}_k$ and $\mathbf{x}_{k+t}$ are the network states at time step $k$ and $k + t$. We can frame this as a trajectory prediction problem. For a given state $\mathbf{x}_k$, we need to minimise the differences between the actual $\mathbf{x}_{k+t} = f^t(\mathbf{x}_k)$ and the predicted $\Hat{\mathbf{x}}_{k+t} = \varphivec^{-1} (\mathbf{K}^t (\varphivec(\mathbf{x}_k)))$. In other words, we are trying to find a latent space representation of the dynamics such that one time step corresponds to one linear transformation in the latent space. $\varphivec$ encodes network states into their latent embeddings, $\mathbf{K}$ linearly advances the latent embeddings, and $\varphivec^{-1}$ decodes latent embeddings back into the original dynamics.

\newcommand{\recon}{\mathcal{L}_{recon}}
\newcommand{\lin}{\mathcal{L}_{linear}}
\newcommand{\pred}{\mathcal{L}_{pred}}

We train our models on three types of loss functions \cite{lusch_kutz_brunton_2018}. The first type is the reconstruction loss:
\[\recon = ||\ \mathbf{x} - \varphivec^{-1}(\varphivec(\mathbf{x}))\ ||\]
which ensures the network dynamics are uniquely recoverable from the latent embeddings. The second type is the linear prediction loss:
\[\lin = ||\ \varphivec(\mathbf{x}_{k+t}) - \mathbf{K}^t(\varphivec(\mathbf{x}_k))\ ||\]
which enforces the latent space structure such that latent embeddings advance linearly with $\mathbf{K}$. The third type is the dynamics prediction loss:
\[\pred = ||\ \mathbf{x}_{k+t} - \varphivec^{-1}(\mathbf{K}^t(\varphivec(\mathbf{x}_k)))\ || \]
This component ensures the decoder $\varphivec^{-1}$ works with the structure of the latent space to predict future states. $||\cdot||$ refers to the mean-squared error, averaged over time steps and the number of samples. The overall loss function is a linear combination of the three, with weights $\alpha_1, \alpha_2, \alpha_3 \in \mathbb{R}^+$:
\[\mathcal{L} = \alpha_1\recon + \alpha_2\lin + \alpha_3\pred\]

\subsection{KMPNN Autoencoder}

Let $G = (V, E)$ be a graph with $n$ nodes, $m$ edges, and an adjacency matrix $\mathbf{A}\in \mathbb{R}^{n\times n}$. The value of node $u$ at time step $t$ in the network dynamics is given by $v_{u, t}$. In our approach, an embedding $\mathbf{x}_{u, t}$ for each node $u$ and time step $t$ is constructed:
\[\mathbf{x}_{u, t} = \sigma_N(id_u)\ ||\ w(v_{u, t})\]
where $\sigma_N$ is a trainable lookup table encoding node identifiers, and $w$ is an MLP encoding node values, with $||$ denoting concatenation. $\sigma_N(id_u)$ is node/neighbourhood-specific, while $w(v_{u, t})$ depends solely on the current state of a  node. Concatenating these representations gives an embedding for the node. The embedding of an edge $u\to v\in E$ is constructed in a similar way:
\[\mathbf{e}_{u, v} = \sigma_E(id_{u\to v}) \]
with $\sigma_E$ being a trainable lookup table encoding edge identifiers. These higher-dimensional node and edge embeddings are created to enrich the feature space of the network.

Denote the node embedding matrix $\mathbf{X}$ and edge embedding matrix $\mathbf{E}$ by:
\begin{align*}
    & \mathbf{X} = \begin{bmatrix} \mathbf{x}_{1} \mathbf{x}_{2} \dots \mathbf{x}_{n} \end{bmatrix}^{\intercal} \in \mathbb{R}^{n\times c}    \\
    & \mathbf{E} = \begin{bmatrix} \mathbf{e}_{u_1,v_1} \mathbf{e}_{u_2, v_2} \dots \mathbf{e}_{u_m, v_m} \end{bmatrix}^{\intercal} \in \mathbb{R}^{m\times c}
\end{align*}
with $c$ being the dimension of the trainable lookup tables. Apply two MPNN layers $\Theta_{\mathbf{A}}$ and $\Theta_{\mathbf{A}}^{(1)}$ on node embeddings $\mathbf{X}$ and $\mathbf{E}$:
\begin{align*}
    \mathbf{X}^{(1)} &= \Theta_{\mathbf{A}}(\mathbf{X}, \mathbf{E}) \\
    \mathbf{X}^{(2)} &= \Theta_{\mathbf{A}}^{(1)}(\mathbf{X}^{(1)}, \mathbf{E}) \\
    \mathbf{y} &= \phi(\mathbf{X}^{(2)})
\end{align*}
where $\phi: \mathbb{R}^{n\times c}\to \mathbb{R}^h$ is an MLP with $h$ being the dimension of the linear latent space. Here we use MPNN layers over other graph convolutions for their expressiveness, whereas the MLP provides the flexibility to learn the message generation and node update functions. This defines the graph encoder $\varphivec$ in our approach.

The latent representations after $t$ time steps are given by:
\[\mathbf{\widetilde{y}} = \mathbf{K}^{t}\mathbf{\dot{y}}\]
where $\mathbf{K}\in \mathbb{C}^{0.5h\times 0.5h}$ is a diagonal matrix and $\mathbf{\dot{y}} \in \mathbb{C}^{0.5h}$. The vector $\mathbf{\dot{y}}$ is a regrouping of $\mathbf{y} \in \mathbb{R}^h$ into pairs that represent the real and imaginary components of complex numbers. In a slight abuse of notation, the resulting product is subsequently arranged back into a real vector. Complex eigenvalues are used in the diagonal matrix $\mathbf{K}$ to accommodate periodic orbits in the dynamics. For ease of reading, we omit the details of these operations from the equation above.

The graph decoder $\varphivec^{-1}$ can be defined in a similar way:
\begin{align*}
    \mathbf{\widetilde{X}}^{(2)} &= 
        \phi^{-1}\left(\mathbf{\widetilde{y}}\right) \\
    \mathbf{\widetilde{X}}^{(1)} &= 
        \Theta_{\mathbf{A}}^{(1)'} \left(
        \mathbf{\widetilde{X}}^{(2)},
        \mathbf{E}\right) \\
    \mathbf{\widetilde{X}} &= 
        \Theta_{\mathbf{A}}' \left({\mathbf{\widetilde{X}}^{(1)}, \mathbf{E}}\right)
\end{align*}
with $\phi^{-1}: \mathbb{R}^{h}\to \mathbb{R}^{n\times c}$ being an MLP as well. Finally, a per-node MLP is applied to the activation of every node to obtain a prediction value for each node representing its current state.

The KMPNN autoencoder exploits the ``neighbourhood" concept in the geometry of the dynamics. Each graph convolution layer uses MLPs for flexible message generation and node updates. In the graph encoder, the  last-layer activations are concatenated and fed into a global MLP to create an overall latent embedding. This step breaks the permutation-invariant property of graph neural networks in general. Still, since the dynamics have the same geometry across different trajectory runs, the property plays a less significant role in achieving our aim.

\section{Experiments}
\label{sec:experiments}

In this section, we evaluate our approach by measuring its performance in predicting the trajectories of several network dynamical systems under random initial states and comparing it against conventional Koopman analysis baselines.

\subsection{Experimental Setup}

\subsubsection{Datasets}

We evaluate our approach on two types of dynamics: (1) synthetic network dynamics governed by specific differential equations and (2) training dynamics of neural networks. The synthetic dynamics are generated by following through the equations from random initial states, which can be found in \cref{table:synthetic_network_dynamics}. These dynamics are modelled as discrete-time dynamical systems by specifying a $\Delta t$ per snapshot and approximating the following snapshot with the composite trapezoidal rule. 

The weight/bias trajectories of neural network training are created by applying mini-batch gradient descent on randomly-initialised neural networks, with different non-trivial architectures, on a suite of tasks. Details about these tasks are in \cref{appendix:training_dynamics_specs}. After running through the entire training dataset, the updated set of weights and biases becomes the state of the next epoch. Note that neural network training dynamics are intrinsically discrete-time dynamical systems.

We generate 10k trajectories for each dynamical system, with train, validation, and test splits of 8k, 1k, and 1k, respectively. All initial states in the synthetic dynamics are uniformly sampled within $[0, 1]$, while the initial weights and biases in the training dynamics are sampled in $[-1, 0.1]$. The number of snapshots for each dynamical system is carefully chosen to ensure convergence whenever possible, usually between 50-500 time steps. The graphs in the synthetic networks all consist of 100 nodes and 250 bidirectional edges, while the architectures of the neural networks range between 50-300 parameters.


\subsubsection{Baselines}

We compare our approach against four data-driven baselines: dynamic mode decomposition (DMD) \cite{H_Tu_2014}, Extended DMD \cite{folkestad_2019}, Graph DMD \cite{fujii_2019}, and the Lusch autoencoder \cite{lusch_kutz_brunton_2018}. DMD is a classical algorithm \cite{schmid2010dynamic} that identifies a best-fit linear approximation to the Koopman operator of a dynamical system by seeking the dominant eigenvalues and eigenvectors in the spectral decomposition of the linear operator \cite{rowley2009spectral}. Extended DMD augments the identity measurements in DMD by non-linear measurements before performing the regression \cite{williams_2015}. Graph DMD is a state-of-the-art approach that extends DMD to graph dynamics by utilising tensor-train decomposition to maintain the tensor structure of the dynamics \cite{fujii_2019}. The Lusch autoencoder uses feedforward MLPs to approximate a set of measurement functions that advance linearly upon being acted on by the Koopman operator, such that the measurements produced uniquely represent the original dynamics \cite{lusch_kutz_brunton_2018}. To ensure a fair comparison, the Lusch autoencoder used in the experiments has a similar number of parameters as our proposed KMPNN autoencoder.

\subsubsection{Training}

The parameters of the deep learning models were initialised uniformly random in the range $(-\frac{1}{\sqrt{d}}, \frac{1}{\sqrt{d}})$ with $d$ being the input dimension of their layers. The Adam optimiser used had a  default learning rate of $10^{-3}$ for optimisation. The models were trained on an NVIDIA Titan Xp GPU, with each Lusch autoencoder taking 3-12 hours and the KMPNN autoencoder taking 8-24 hours. 

The Lusch autoencoder uses a 3-layer MLP for both its encoder and decoder. Unless otherwise specified, both autoencoders are trained with a latent space dimension being the closest power of 2 greater than the number of nodes in the dynamics.

\subsection{Synthetic Network Dynamics}

\begin{table*}[t]
\setlength\tabcolsep{1.5pt}
\caption{Prediction losses of our approach on synthetic network dynamics. The Lusch and KMPNN autoencoders are trained with a latent space dimension of 256. All entries for extended DMD and some for graph DMD are omitted as the algorithm is unable to identify a system, or the identified system produced out-of-scale predictions.} 
\label{table:synthetic_prediction_losses}
\vskip 0.1in
\begin{center}
\begin{small}
\begin{sc}
\resizebox{\columnwidth}{!}{\begin{tabular}{@{\extracolsep{3pt}}*{6}{c}}
    \toprule
    Baseline                & \textbf{Regulatory}   & \textbf{Neuronal}     & \textbf{Population}   & \textbf{Epidemic}     & \textbf{Mutualistic}  \\
    \midrule
    DMD                     & 0.0726                & 0.8482                & 1.0638                & 0.0175                & 0.3891                \\
    Graph DMD               & 0.4443                & -                     & -                     & 0.0465                & 1.1023                \\
    Lusch autoencoder       & 0.0340                & 0.0282                & 0.0484                & 2.576e-3                & 0.0427                \\
    KMPNN autoencoder       & \textbf{4.935e-5}     & \textbf{4.716e-5}     & \textbf{1.042e-4}         & \textbf{8.142e-6}     & \textbf{3.775e-4}       \\
     \bottomrule
\end{tabular}}
\end{sc}
\end{small}
\end{center}

\vskip -0.1in
\end{table*}

\cref{table:synthetic_prediction_losses} shows the prediction losses of the models in the trajectories of the synthetic dynamics based on differential equations from a variety of disciplines \cite{bakhtiarnia_2020}. The prediction loss can be interpreted as the average squared deviation of each node between the predicted and actual trajectories. Our model significantly outperforms existing baselines in capturing a set of measurement functions and a linear Koopman operator to form a Koopman-invariant subspace. This is reflected by the better prediction performance with the linear representation captured by the measurement functions in our approach.

In our experiments, the deep learning approaches show a substantial advantage over theoretical baselines. DMD and Graph DMD fit a linear model to the dynamics, so their inability to capture non-linearity could be expected. Extended DMD assumes that a Koopman-invariant subspace can be approximated with a library of candidate functions, relying on the library is sufficiently expressive to represent the dynamics. When this is not the case, the “invariant” system identified would be largely spurious, accounting for its inability to produce sound predictions.

All baselines, except for Graph DMD, do not consider the inherent graph structures of the dynamics. The baselines' expressiveness could be constrained without considering the dynamics' structure, especially in the Lusch autoencoder. Without the inductive bias, the Lusch autoencoder flattens the graph dynamics before feeding into the MLPs. Being quadratic to the input dimension, the fully-connected layers could impede the search for good minima during optimisation.

\begin{figure*}
\vskip 0.1in
\begin{center}
\caption{Example of synthetic network dynamic trajectories and the corresponding predictions made by DMD (purple dashed line), Graph DMD (orange dashed line), Lusch autoencoder (green dashed line), and KMPNN autoencoder (red dashed line). The x-axis and y-axis each represents a node.}
\label{fig:node_trajs}
\subfigure[Epidemic dynamics]{\includegraphics[scale=0.35]{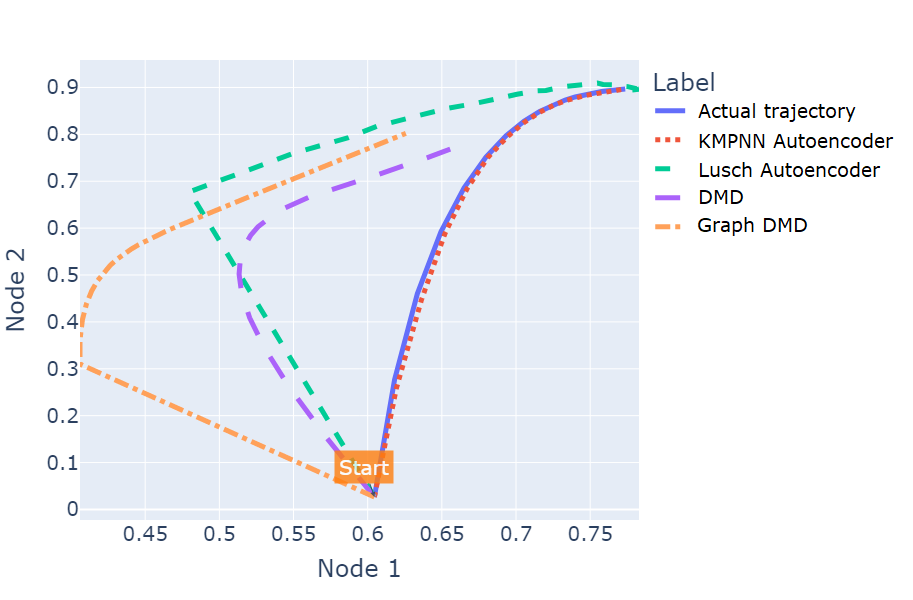}}
\subfigure[Mutualistic dynamics]{\includegraphics[scale=0.35]{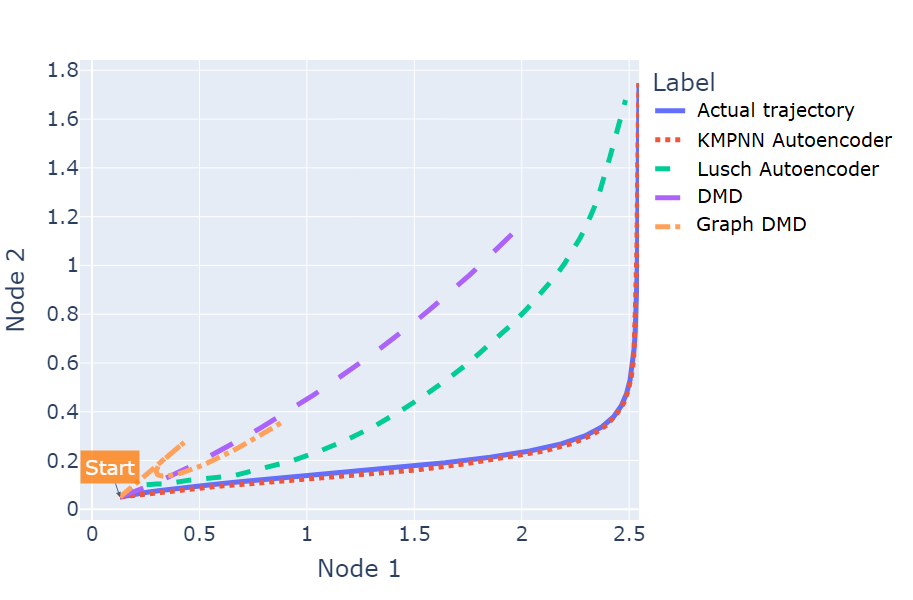}}
\end{center}
\vskip -0.1in
\end{figure*}

\cref{fig:node_trajs} shows some qualitative results comparing the actual and predicted trajectories of two nodes in the test samples. The KMPNN approach accurately predicts the trajectories from unseen starting conditions. Most baselines predict similarly well for population dynamics, likely because many of the trajectories are close to being linear and do not converge. In contrast, predictions in epidemic and mutualistic dynamics make the differences in predictive power between the theoretical baselines, the Lusch autoencoder, and our approach more evident. The prediction of our approach closely resembles the actual trajectories, while other baselines are less successful in capturing the dynamics, despite being able to arrive at a similar state eventually.

\subsection{Neural Network Training Dynamics}

\begin{table*}[t]
\setlength\tabcolsep{1.5pt}
\caption{Prediction losses of our approach on neural network training dynamics. The architectures of the neural networks (whose training dynamics are captured) are labelled with either $x$-fc or $y$-conv-$z$-fc. The former refers to an architecture with $x$ fully-connected layers, the latter with $y$ sets of convolutional and max pooling layers followed by $z$ fully-connected layers. Some extended and Graph DMD entries are omitted as the algorithm is unable to identify a system, or the identified system produced out-of-scale predictions.}
\label{prediction losses}
\vskip 0.1in
\begin{center}
\begin{small}
\begin{sc}
\resizebox{\columnwidth}{!}{\begin{tabular}{@{\extracolsep{3pt}}*{7}{c}}
    \toprule
        & \textbf{Linear} & \multicolumn{2}{c}{\textbf{Classification}} & \textbf{DE solver} & \multicolumn{2}{c}{\textbf{Image classification}} \\
    \cline{2-2}\cline{3-4}\cline{5-5}\cline{6-7}
        Baseline            & 1-fc              & 2-fc              & 3-fc              & 3-fc              & 1-conv-2-fc       & 2-conv-1-fc \\
    \midrule
    DMD                     & 48.1506           & 0.0591            & 0.0886            & 0.0671            & 0.4245            & 0.3687 \\
    Extended DMD            & -                 & 0.0710            & -                 & 0.1001            & -                 & - \\
    Graph DMD               & -                 & 0.0560            & 0.0954            & 0.0846            & -                 & - \\
    Lusch autoencoder       & 48.1532           & 0.0020            & 0.0305            & 0.0323            & 0.1203            & 0.2463 \\
    KMPNN autoencoder         & \textbf{48.1291}  & \textbf{0.0006}   & \textbf{0.0276}   & \textbf{0.0119}   & \textbf{0.0366}   & \textbf{0.1153} \\
     \bottomrule
\end{tabular}}
\end{sc}
\end{small}
\end{center}
\vskip -0.1in
\end{table*}

\begin{figure*}
\vskip 0.1in
\begin{center}
\caption{Example of weight and bias trajectories and the corresponding predictions made by DMD (purple dashed line), Lusch autoencoder (green dashed line), and KMPNN autoencoder (red dashed line). The x-axis and y-axis each represents a weight/bias.}
\label{fig:weight_bias_trajs}
\subfigure[DE solver: 3-fc]{\includegraphics[scale=0.35]{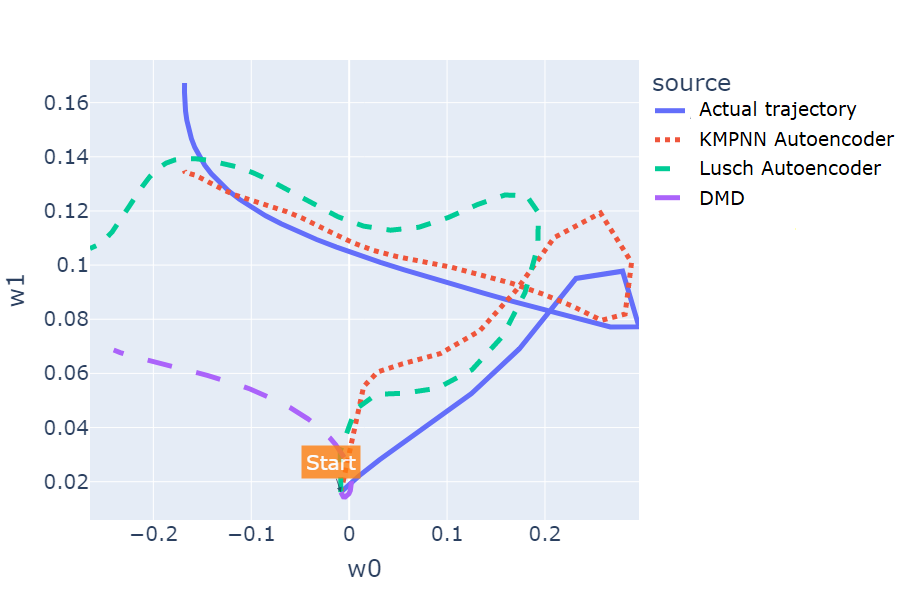}}
\subfigure[Image classification: 1-conv-2-fc]{\includegraphics[scale=0.35]{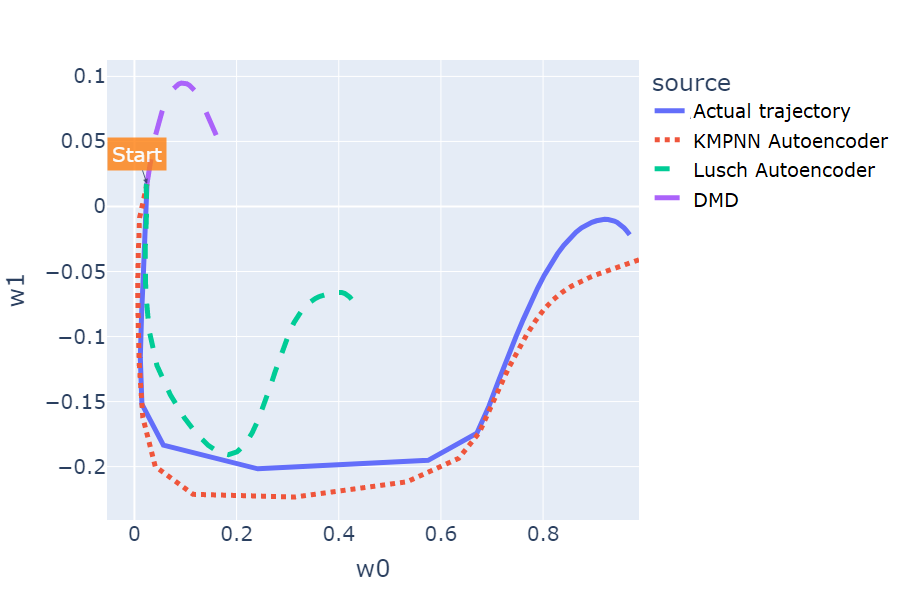}}
\end{center}
\vskip -0.1in
\end{figure*}

\cref{prediction losses} shows the prediction losses of models in weight and bias training dynamics of different neural networks. The difference in predictive abilities of our approach and the Lusch autoencoder becomes much more evident (except for the linear training trajectories for linear regression, which linear approaches like DMD can already successfully capture). Backpropagation in gradient descent stacks non-linear gradient terms one-within-another, exponentially complexifying the gradient expressions. Since the non-linearity in training dynamics requires a more complex set of measurement functions, this can explain the widened performance difference between our approach, the Lusch autoencoder, and other baselines. \cref{fig:weight_bias_trajs} plots the predicted trajectories of two weights/biases against their actual trajectory.

\subsubsection{Optimisation Performance}

\begin{figure*}
\vskip 0.1in
\begin{center}
\caption{Example of actual and predicted loss trajectories made by DMD (purple dashed line), Lusch autoencoder (green dashed line), and KMPNN autoencoder (red dashed line).}
\label{fig:loss_trajectories}
\subfigure[DE solver: 3-fc]{\includegraphics[scale=0.35]{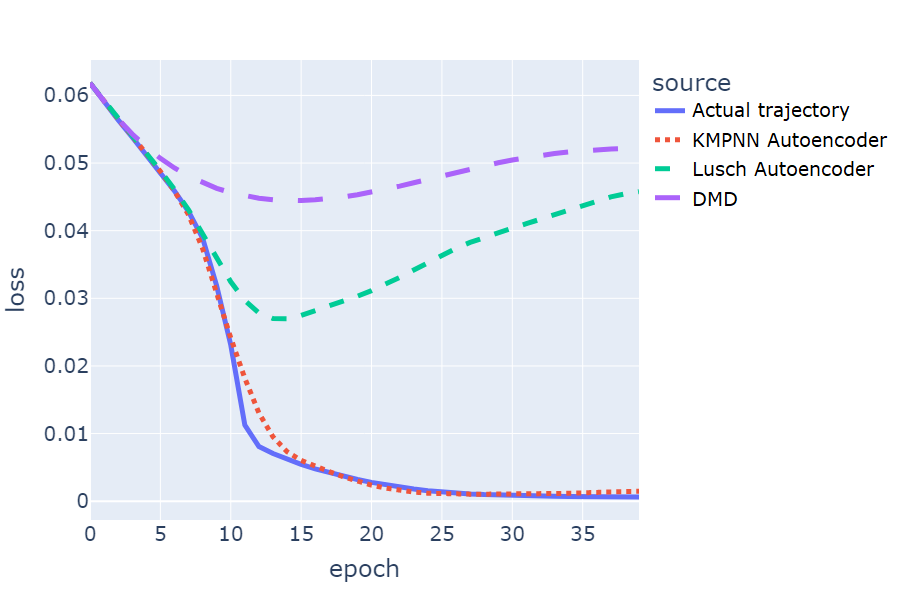}}\quad
\subfigure[Image classification: 1-conv-2-fc]{\includegraphics[scale=0.35]{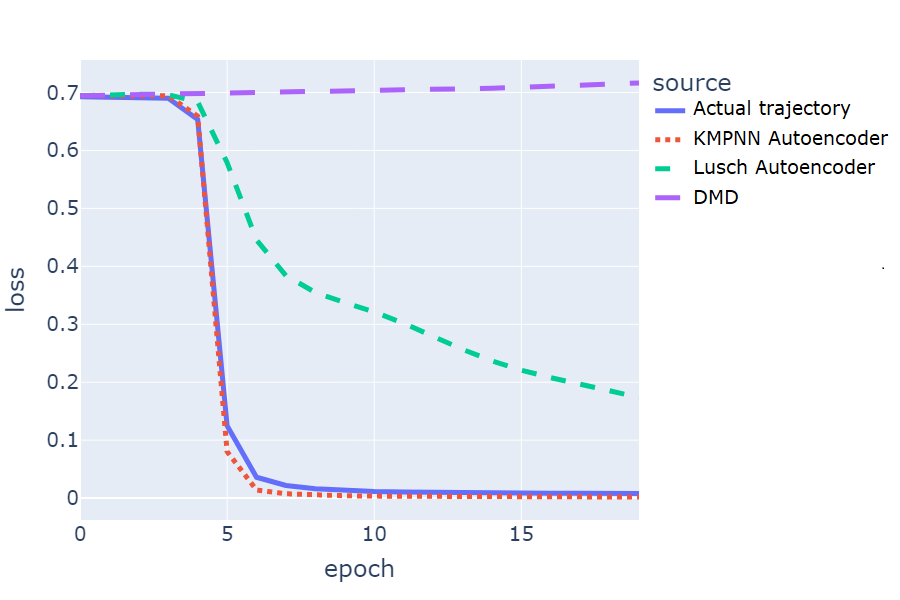}}
\end{center}
\vskip -0.1in
\end{figure*}

We look into the loss trajectories of the neural networks using the weights and biases predicted by the models. \cref{fig:loss_trajectories} shows a qualitative comparison between the actual loss trajectories and the loss trajectories using predicted weights and biases. Despite being unable to predict the exact trajectory on which the weights and biases evolve, our approach can find a trajectory with a similar loss (if not better) performance that can be represented linearly in the Koopman-invariant subspace. The predicted trajectories often avoid training instability, converging along a smoother path.

A similar evaluation methodology as \citet{dogra_redman_2020} is adopted to quantify the optimisation performance of our approach. We look at the optimisation performance $r$, the ratio of the differences between the initial loss and the final loss in the actual trajectory to that of the predicted trajectory:
\[r = \frac{l_0 - l_{pred}}{l_0 - l_{true}}\]
with $l_0$ being the initial training loss, and $l_{pred}$ and $l_{true}$ being the final training losses in the predicted and actual trajectories, respectively. The closer $r$ is to 1, the closer the final loss of the predicted trajectory is to the actual trajectory. The average values for $r$ (in percentages) are tabulated in \cref{table:optimisation_performances}, which shows the weight and bias trajectories predicted using our approach indeed have a similar optimisation performance as backpropagation.

\begin{table*}[t]
\setlength\tabcolsep{1.5pt}
\caption{Optimisation performances of weights and biases predicted with our approach.}
\label{table:optimisation_performances}
\vskip 0.1in
\begin{center}
\begin{small}
\begin{sc}
\begin{tabular}{@{\extracolsep{3pt}}*{6}{c}}
    \toprule
        \textbf{Linear} & \multicolumn{2}{c}{\textbf{Classification}} & \textbf{DE solver} & \multicolumn{2}{c}{\textbf{Image classification}} \\
    \cline{1-1}\cline{2-3}\cline{4-4}\cline{5-6}
        1-fc              & 2-fc              & 3-fc              & 3-fc              & 1-conv-2-fc       & 2-conv-1-fc \\
    \midrule
        100.3\%           & 98.6\%            & 92.2\%            & 87.6\%            & 100.6\%           & 90.5\% \\
     \bottomrule
\end{tabular}
\end{sc}
\end{small}
\end{center}
\vskip -0.1in
\end{table*}

\subsubsection{Latent Space Dimensions}

An interesting observation comes from analysing how many latent dimensions are required for the measurement functions to capture the training dynamics. The dynamics of two neural networks are studied, including a 2-layer MLP on the classification task (with 106 parameters) and a 3-layer CNN on the image classification task (with 194 parameters). \cref{table: latent_space_dimensions} reports the prediction losses and optimisation performances of the KMPNN autoencoder equipped with a count of measurement functions varying along powers of 2. 

Our approach can learn a representation for the training dynamics of neural networks with $>100$ parameters, starting from random initialisations, using $<10$ linearly-evolving eigenfunctions. The possibility of a low-dimensional linear representation provides another perspective for the general over-parametrisation phenomena of neural networks and is in line with work studying the dynamics present during deep neural network training \cite{gur2018gradient, turjeman2022underlying}.

\begin{table}[t]
\caption{Prediction losses and optimisation performances $r$ (in percentages) of our approach evaluated on the training dynamics of a 2-fc MLP and a 1-conv-2-fc CNN with latent dimension varying among powers of 2.}
\label{table: latent_space_dimensions}
\vskip 0.15in
\begin{center}
\begin{small}
\begin{sc}
\begin{tabular}{lcccr}
\toprule
    Dimension & 2-fc MLP & 1-conv-2-fc CNN \\
\midrule
    4   &   0.0130 / 90.1\%  & 0.4536 / 0.1\%     \\
    8   &   0.0030 / 98.2\%  & 0.0562 / 100.7\%   \\
    16  &   0.0022 / 98.5\%  & 0.0647 / 100.6\%   \\
    32  &   0.0015 / 98.5\%  & 0.0805 / 100.5\%   \\
    64  &   0.0006 / 98.4\%  & 0.0513 / 100.7\%   \\
    128 &   0.0005 / 98.0\%  & 0.0393 / 100.6\%   \\
\bottomrule
\end{tabular}
\end{sc}
\end{small}
\end{center}
\vskip -0.1in
\end{table}

\subsubsection{Robustness to Variations}

We show the robustness of our KMPNN autoencoder by evaluating it on training dynamics with varying specifications. Our approach maintains a similar optimisation performance when applied to the training dynamics of neural networks with different non-linearity activation functions, optimisers, and randomised mini-batches. More details can be found in \cref{appendix:robustness}.

\subsubsection{Comparison with Previous Approaches}

There have been several approaches \cite{dogra_redman_2020,tano_2020} attempting to identify a linear relationship on which the weights and biases evolve. The linear Koopman operators identified by both papers are local to a single perceptron/layer at a specific training instance. The operator \citet{tano_2020} identified only applies to a few training epochs ahead, while the one by \citet{dogra_redman_2020} is valid only after training has entered basins of local loss minima. 

Our approach suffers from none of these limitations. The KMPNN autoencoder can identify a globally linear operator (and a set of measurement functions) representing the entire neural network's dynamics. The same operator can apply to any randomly initialised trajectory. The operator identified is also valid from the beginning to the end of the training process without requiring proximity to any loss minima.

\section{Conclusion}

We have presented a novel, highly effective approach for finding a linear representation of non-linear network dynamics. We successfully demonstrated that the linear representation produced by our approach provides better predictions of the dynamics than the state-of-the-art. Although our approach takes more wall clock time to train, it significantly outperforms existing traditional and deep learning approaches for non-linear dynamics problems, including neural network training dynamics. In the latter problem, the linear representations extracted are significantly smaller than the respective network sizes.

The optimisation performance of the weights and biases predicted in neural network training are on par with classical optimisers while robust to architectural and training configuration variations. The linearisation is also more versatile than previous approaches \cite{dogra_redman_2020, tano_2020}. 

The KMPNN autoencoder presented in this paper has significant potential for application to other nonlinear network dynamics problems. We hope this success will lead future efforts to address further challenges in the application of Koopman operatory theory to complex non-linear systems. For example, as neural networks tend to have many parameters, each of which our approach treats as a node, the KMPNN is limited by the amount of GPU memory available when applied to larger and denser architectures. A direction for future research could be to scale our approach to bigger networks by reducing the memory footprint of the KMPNN autoencoder.

\bibliography{main}

\newpage
\appendix
\onecolumn

\section{Datasets of Neural Network Training Trajectories}

\label{appendix:training_dynamics_specs}

Different tasks are formulated to study the dynamics prediction models' abilities and limitations. Neural networks of various depths, equipped with different types of non-linearity, are trained to tackle the tasks. Trajectory datasets are created by training the neural networks with random initial conditions for a fixed number of epochs. After each epoch, weights, biases, and training losses are collected to form the training trajectories.

The following is the list of tasks studied in the experiments:

\textbf{Linear regression}\quad The linear regression problem is based on a synthetic dataset generated using the formula:
\[y_i = \mathbf{w}\cdot \mathbf{x}_i + b + \epsilon_i\]
with $\epsilon_i$ sampled from a Gaussian distribution. 20 features are synthesied, and 2000 sampels are collected. \textbf{1-layer} feedforward network is trained for this problem.

\textbf{Classification}\quad The task is based on the classic wine dataset, which results from a chemical analysis of wines grown in the same region in Italy but derived from 3 different cultivars \cite{uci_wine_dataset}. The task is to classify each wine into 1 of the 3 classes using the quantities of its 13 constituents. The dataset consists of 178 samples.

Two feedforward NNs are trained for this problem:
\begin{itemize}
    \item \textbf{2-layer}\quad Linear $\rightarrow$ Activation $\rightarrow$ Linear $\rightarrow$ Softmax
    \item \textbf{3-layer}\quad Linear $\rightarrow$ Activation $\rightarrow$ Linear $\rightarrow$ Activation $\rightarrow$ Linear $\rightarrow$ Softmax
\end{itemize}

\textbf{Image classification}\quad The task is based on the digits dataset, which includes 1800 handwritten digits in 8x8 bitmaps \cite{uci_digits_dataset}. The objective is to classify each bitmap as 1 of the 10 digits.

This task aims to study the models' performance in predicting the dynamics of convolutional neural networks (CNN). To reduce memory usage, only images from 2 classes are included so that the CNN can be slightly scaled down.

Define a \textit{convolutional layer} as "Convolution $\rightarrow$ Activation $\rightarrow$ Average Pooling". Two CNNs are trained for this problem:
\begin{itemize}
    \item \textbf{1-conv 2-fc CNN}\quad Convolutional layer $\rightarrow$ Linear $\rightarrow$ Activation $\rightarrow$ Linear
    \item \textbf{2-conv 1-fc CNN}\quad Convolutional layer $\rightarrow$ Convolutional layer $\rightarrow$ Linear
\end{itemize}

\textbf{DE solver}\quad A neural network differential equation (DE) solver is trained to solve for $f(x)$, where:
\[f'(x) - f(x) = y\]
with $f(1) = 0$ as the initial conditions.

The DE solver in the experiment comprises 3 layers, each separated by a non-linear activation layer. The inputs are uniformly sampled over a range on which the loss is evaluated.

The tasks are selected to include both synthetic and real-world problems. Input normalisation (i.e., standardising each dimension to a mean of 0 and variance of 1) is applied to some task datasets to make the weights and biases evolve within a narrower range in training.

The initial weights and biases are sampled from $(-1, 1)$. The neural networks are then trained with mini-batch gradient descent so that the training process corresponds precisely to the mathematical framework outlined in the last chapter. To avoid complex dynamics (e.g., involving momentum), the SGD (Stochastic Gradient Descent) optimiser is used in the study. In \cref{appendix:robustness}, some of these configurations are altered to analyse the models' robustness to variations.

To keep the training time of baselines within a reasonable limit, the weights and biases are sampled per 10 epochs to shorten the trajectories. This augmentation would have a minimal effect on the predictions, as the objective of the predictions is to capture the overall trends instead of transient fluctuations.
\clearpage

\section{Robustness to Architectural and Training Variations}
\label{appendix:robustness}

This section aims to showcase the robustness of our KMPNN autoencoder when confronted with training trajectories with various specifications.

\subsection{Non-linearity}

To investigate the versatility of our KMPNN autoencoder on different non-linearity types, the model is evaluated on the training trajectories of the same neural architectures equipped with different non-linear activation functions. The model is again evaluated on the dynamics of a 2-layer MLP and a 3-layer CNN. The optimisation performances $r$ are as follows:

\begin{table*}[h!]
\setlength\tabcolsep{1.5pt}
\vskip 0.1in
\begin{center}
\begin{small}
\begin{sc}
\begin{tabular}{@{\extracolsep{3pt}}*{6}{c}}
    \toprule
        & \textbf{ELU ($\alpha=1$)}  & \textbf{ReLU}     & \textbf{Leaky ReLU}   & \textbf{Sigmoid}  & \textbf{Tanh} \\
    \midrule
    Classification: 2-fc & 98.6\% & 97.3\% & 96.7\% & 99.2\% & 98.5\% \\
    Image: 1-conv-2-fc & 96.9\% & 83.6\% & 93.5\% & 100.6\% & 98.0\% \\
    \bottomrule
\end{tabular}
\end{sc}
\end{small}
\end{center}
\vskip -0.1in
\end{table*}

In general, the losses of neural architectures equipped with sigmoid, tanh, and ELU are better predicted than ReLU and leaky ReLU. This might suggest that smoothness of activation functions play a role here since sigmoid, tanh, and ELU (with $\alpha=1$) are smooth functions, while the gradients of ReLU and leaky ReLU are discontinuous at $x=0$. Nevertheless, the loss performances are robust regardless of the choice of activation functions.

\subsection{Optimisers}

In this section, our approach is evaluated on the training trajectories produced by 4 different optimisers, which include SGD, Adagrad \cite{adagrad_2011}, Adadelta \cite{adadelta_2012} and Adam \cite{adam_2014}. The optimisation performances are:

\begin{table*}[h!]
\setlength\tabcolsep{1.5pt}
\vskip 0.1in
\begin{center}
\begin{small}
\begin{sc}
\begin{tabular}{@{\extracolsep{3pt}}*{5}{c}}
    \toprule
        & \textbf{SGD}    & \textbf{Adagrad}   & \textbf{Adadelta}   & \textbf{Adam} \\
    \midrule
    Classification: 2-fc & 98.6\% & 98.1\% & 94.2\% & 98.8\% \\
    Image: 1-conv-2-fc & 100.6\% & 95.7\% & 97.7\% & 99.2\% \\
    \bottomrule
\end{tabular}
\end{sc}
\end{small}
\end{center}
\vskip -0.1in
\end{table*}

As shown, our approach is robust across the choice of optimisers.

\subsection{Stochastic Training}

Recall that the training trajectories used throughout the experiments are generated with mini-batch gradient descent, i.e., training data are grouped into small batches for multiple passes in an epoch. However, most training pipelines now use shuffled batches to improve generalisation. To show that our approach is robust against variance among random mini-batches, this subsection reports the optimisation performances of our KMPNN autoencoder when applied to stochastic trajectories:

\begin{table*}[h!]
\setlength\tabcolsep{1.5pt}
\vskip 0.1in
\begin{center}
\begin{small}
\begin{sc}
\begin{tabular}{@{\extracolsep{3pt}}*{5}{c}}
    \toprule
         & \textbf{SGD}    & \textbf{Adagrad}   & \textbf{Adadelta}   & \textbf{Adam} \\
    \midrule
    Classification: 2-fc & 99.9\% & 97.4\% & 99.7\% & 99.6\% \\
    Image: 1-conv-2-fc & 97.8\% & 99.4\% & 99.3\% & 99.1\% \\
    \bottomrule
\end{tabular}
\end{sc}
\end{small}
\end{center}
\vskip -0.1in
\end{table*}

The results show that our approach can withstand variances among random mini-batches in training data.
\clearpage


\end{document}